\documentclass[a4paper,UKenglish,cleveref,autoref,thm-restate]{lipics-v2021}

\pdfoutput=1

\usepackage{amssymb}
\usepackage{booktabs}

\newcommand{\RR}{\mathbb{R}}
\newcommand{\NN}{\mathbb{N}}

\newcommand{\eps}{\varepsilon}

\newcommand{\calX}{\mathcal{X}}
\newcommand{\calF}{\mathcal{F}}

\newcommand{\emprad}{\widehat{\mathfrak{R}}}
\newcommand{\rad}{\mathfrak{R}}

\newcommand{\emb}[1]{\textbf{\textcolor{blue}{#1}}}
 
\usepackage[T1]{fontenc}
\usepackage[utf8]{inputenc}
\usepackage[scaled=0.9]{DejaVuSansMono}    \usepackage{listings}

\usepackage{xcolor}
\definecolor{keywordcolor}{HTML}{4169e1}   \definecolor{tacticcolor}{HTML}{4169e1}    \definecolor{commentcolor}{HTML}{2e8b57}   \definecolor{symbolcolor}{HTML}{000000}\definecolor{sortcolor}{HTML}{4169e1}      \definecolor{attributecolor}{HTML}{f75394} \definecolor{bgcolor}{gray}{0.95}

\lstdefinelanguage{lean} {
mathescape=false,
texcl=false,
morekeywords=[1]{
import, prelude, protected, private, noncomputable, definition, meta, renaming,
hiding, parameter, parameters, begin, constant, constants,
lemma, variable, variables, theory,
print, theorem, example,
open, as, export, override, axiom, axioms, inductive, with,
structure, record, universe, universes,
alias, help, precedence, reserve, declare_trace, add_key_equivalence,
match, infix, infixl, infixr, notation, postfix, prefix, instance,
eval, reduce, check, end, this,
using, using_well_founded, namespace, section,
attribute, local, set_option, extends, include, omit, class,
raw, replacing,
calc, have, show, suffices, by, in, at, let, forall, Pi, fun,
exists, if, dif, then, else, assume, obtain, from, register_simp_ext, unless, break, continue,
mutual, do, def, run_cmd, const,
partial, mut, where, macro, syntax, deriving,
return, try, catch, for, macro_rules, declare_syntax_cat, abbrev,
ring,
rcases,
no, goals,
},
morekeywords=[2]{Sort, Type, Prop},
morekeywords=[3]{
assumption,
apply, intro, intros, allGoals,
generalize, clear, revert, done, exact,
refine, repeat, cases, rewrite, rw,
simp, simp_all, contradiction,
constructor, injection,
induction,
},
literate=
{α}{{\ensuremath{\mathrm{\alpha}}}}1
{β}{{\ensuremath{\mathrm{\beta}}}}1
{γ}{{\ensuremath{\mathrm{\gamma}}}}1
{δ}{{\ensuremath{\mathrm{\delta}}}}1
{ε}{{\ensuremath{\mathrm{\varepsilon}}}}1
{ζ}{{\ensuremath{\mathrm{\zeta}}}}1
{η}{{\ensuremath{\mathrm{\eta}}}}1
{θ}{{\ensuremath{\mathrm{\theta}}}}1
{ι}{{\ensuremath{\mathrm{\iota}}}}1
{κ}{{\ensuremath{\mathrm{\kappa}}}}1
{μ}{{\ensuremath{\mathrm{\mu}}}}1
{ν}{{\ensuremath{\mathrm{\nu}}}}1
{ξ}{{\ensuremath{\mathrm{\xi}}}}1
{π}{{\ensuremath{\mathrm{\mathnormal{\pi}}}}}1
{ρ}{{\ensuremath{\mathrm{\rho}}}}1
{σ}{{\ensuremath{\mathrm{\sigma}}}}1
{τ}{{\ensuremath{\mathrm{\tau}}}}1
{φ}{{\ensuremath{\mathrm{\varphi}}}}1
{χ}{{\ensuremath{\mathrm{\chi}}}}1
{ψ}{{\ensuremath{\mathrm{\psi}}}}1
{ω}{{\ensuremath{\mathrm{\omega}}}}1
{Γ}{{\ensuremath{\mathrm{\Gamma}}}}1
{Δ}{{\ensuremath{\mathrm{\Delta}}}}1
{Θ}{{\ensuremath{\mathrm{\Theta}}}}1
{Λ}{{\ensuremath{\mathrm{\Lambda}}}}1
{Σ}{{\ensuremath{\mathrm{\Sigma}}}}1
{Φ}{{\ensuremath{\mathrm{\Phi}}}}1
{Ξ}{{\ensuremath{\mathrm{\Xi}}}}1
{Ψ}{{\ensuremath{\mathrm{\Psi}}}}1
{Ω}{{\ensuremath{\mathrm{\Omega}}}}1
{ℵ}{{\ensuremath{\aleph}}}1
{甲}{{\ensuremath{\mathcal{X}}}}1
{≤}{{\ensuremath{\leq}}}1
{≥}{{\ensuremath{\geq}}}1
{≠}{{\ensuremath{\neq}}}1
{≈}{{\ensuremath{\approx}}}1
{≡}{{\ensuremath{\equiv}}}1
{≃}{{\ensuremath{\simeq}}}1
{≤}{{\ensuremath{\leq}}}1
{≥}{{\ensuremath{\geq}}}1
{∂}{{\ensuremath{\partial}}}1
{∆}{{\ensuremath{\triangle}}}1
{∫}{{\ensuremath{\int}}}1
{∑}{{\ensuremath{\mathrm{\Sigma}}}}1
{Π}{{\ensuremath{\mathrm{\Pi}}}}1
{⊥}{{\ensuremath{\perp}}}1
{∞}{{\ensuremath{\infty}}}1
{∂}{{\ensuremath{\partial}}}1
{∓}{{\ensuremath{\mp}}}1
{±}{{\ensuremath{\pm}}}1
{×}{{\ensuremath{\times}}}1
{⊕}{{\ensuremath{\oplus}}}1
{⊗}{{\ensuremath{\otimes}}}1
{⊞}{{\ensuremath{\boxplus}}}1
{∇}{{\ensuremath{\nabla}}}1
{√}{{\ensuremath{\sqrt{}}}}1
{⬝}{{\ensuremath{\cdot}}}1
{•}{{\ensuremath{\cdot}}}1
{∘}{{\ensuremath{\circ}}}1
{⁻}{{\ensuremath{^{-}}}}1
{▸}{{\ensuremath{\blacktriangleright}}}1
{∧}{{\ensuremath{\wedge}}}1
{∨}{{\ensuremath{\vee}}}1
{¬}{{\ensuremath{\neg}}}1
{⊢}{{\ensuremath{\vdash}}}1
{⟨}{{\ensuremath{\langle}}}1
{⟩}{{\ensuremath{\rangle}}}1
{↦}{{\ensuremath{\mapsto}}}1
{←}{{\ensuremath{\leftarrow}}}1
{<-}{{\ensuremath{\leftarrow}}}1
{→}{{\ensuremath{\rightarrow}}}1
{↔}{{\ensuremath{\leftrightarrow}}}1
{⇒}{{\ensuremath{\Rightarrow}}}1
{⟹}{{\ensuremath{\Longrightarrow}}}1
{⇐}{{\ensuremath{\Leftarrow}}}1
{⟸}{{\ensuremath{\Longleftarrow}}}1
{∩}{{\ensuremath{\cap}}}1
{∪}{{\ensuremath{\cup}}}1
{⊂}{{\ensuremath{\subseteq}}}1
{⊆}{{\ensuremath{\subseteq}}}1
{⊄}{{\ensuremath{\nsubseteq}}}1
{⊈}{{\ensuremath{\nsubseteq}}}1
{⊃}{{\ensuremath{\supseteq}}}1
{⊇}{{\ensuremath{\supseteq}}}1
{⊅}{{\ensuremath{\nsupseteq}}}1
{⊉}{{\ensuremath{\nsupseteq}}}1
{∈}{{\ensuremath{\in}}}1
{∉}{{\ensuremath{\notin}}}1
{∋}{{\ensuremath{\ni}}}1
{∌}{{\ensuremath{\notni}}}1
{∅}{{\ensuremath{\emptyset}}}1
{⨆}{{\ensuremath{\sqcup}}}1
{∖}{{\ensuremath{\setminus}}}1
{†}{{\ensuremath{\dag}}}1
{ℕ}{{\ensuremath{\mathbb{N}}}}1
{ℤ}{{\ensuremath{\mathbb{Z}}}}1
{ℝ}{{\ensuremath{\mathbb{R}}}}1
{ℚ}{{\ensuremath{\mathbb{Q}}}}1
{ℂ}{{\ensuremath{\mathbb{C}}}}1
{⌞}{{\ensuremath{\llcorner}}}1
{⌟}{{\ensuremath{\lrcorner}}}1
{⦃}{{\ensuremath{\{\!|}}}1
{⦄}{{\ensuremath{|\!\}}}}1
{⟪}{{\ensuremath{\langle\!\langle}}}1
{⟫}{{\ensuremath{\rangle\!\rangle}}}1
{‖}{{\ensuremath{\|}}}1
{₁}{{\ensuremath{_1}}}1
{₂}{{\ensuremath{_2}}}1
{₃}{{\ensuremath{_3}}}1
{₄}{{\ensuremath{_4}}}1
{₅}{{\ensuremath{_5}}}1
{₆}{{\ensuremath{_6}}}1
{₇}{{\ensuremath{_7}}}1
{₈}{{\ensuremath{_8}}}1
{₉}{{\ensuremath{_9}}}1
{₀}{{\ensuremath{_0}}}1
{ᵢ}{{\ensuremath{_i}}}1
{ⱼ}{{\ensuremath{_j}}}1
{ₐ}{{\ensuremath{_a}}}1
{¹}{{\ensuremath{^1}}}1
{ₙ}{{\ensuremath{_n}}}1
{ₘ}{{\ensuremath{_m}}}1
{ₚ}{{\ensuremath{_p}}}1
{↑}{{\ensuremath{\uparrow}}}1
{↓}{{\ensuremath{\downarrow}}}1
{ⁿ}{{\ensuremath{^n}}}1
{ᵐ}{{\ensuremath{^m}}}1
{...}{{\ensuremath{\ldots}}}1
{·}{{\ensuremath{\cdot}}}1
{▸}{{\ensuremath{\triangleright}}}1
{Σ}{{\color{symbolcolor}\ensuremath{\Sigma}}}1
{Π}{{\color{symbolcolor}\ensuremath{\Pi}}}1
{∀}{{\color{symbolcolor}\ensuremath{\forall}}}1
{∃}{{\color{symbolcolor}\ensuremath{\exists}}}1
{λ}{{\color{symbolcolor}\ensuremath{\mathrm{\lambda}}}}1
{\$}{{\color{symbolcolor}\$}}1
{:=}{{\color{symbolcolor}:=}}1
{=}{{\color{symbolcolor}=}}1
{<|>}{{\color{symbolcolor}<|>}}1
{<\$>}{{\color{symbolcolor}<\$>}}1
{+}{{\color{symbolcolor}+}}1
{*}{{\color{symbolcolor}*}}1,
morecomment=[s][\color{commentcolor}]{/-}{-/},
morecomment=[l][\itshape \color{commentcolor}]{--},
showstringspaces=false,
keepspaces=true,
morestring=[b]",
tabsize=3,
extendedchars=true,
sensitive=true,
breaklines=true,
breakatwhitespace=true,
basicstyle=\ttfamily\small,
captionpos=b,
columns=[l]fullflexible,
identifierstyle={\ttfamily\color{black}},
keywordstyle=[1]{\ttfamily\color{keywordcolor}},
keywordstyle=[2]{\ttfamily\color{sortcolor}},
keywordstyle=[3]{\ttfamily\color{tacticcolor}},
keywordstyle=[4]{\ttfamily\color{attributecolor}},
stringstyle=\ttfamily,
commentstyle={\ttfamily\footnotesize },
}
\title{Lean Formalization of Generalization Error Bound by Rademacher Complexity and Dudley's Entropy Integral}

\titlerunning{Lean Formalization of Rademacher Bounds}

\author{Sho Sonoda\footnote{corresponding author}}{RIKEN AIP, Tokyo, Japan \and CyberAgent Inc., Tokyo, Japan}{sho.sonoda@riken.jp}{https://orcid.org/0000-0001-7242-4740}{}\author{Kazumi Kasaura}{OMRON SINIC X Corporation, Tokyo, Japan}{kazumi.kasaura@sinicx.com}{https://orcid.org/0000-0002-3219-9961}{}\author{Yuma Mizuno}{University College Cork, Cork, Ireland}{mizuno.y.aj@gmail.com}{https://orcid.org/0000-0001-8056-7477}{}
\author{Kei Tsukamoto}{The University of Tokyo, Tokyo, Japan}{milanotukamoto@g.ecc.u-tokyo.ac.jp}{}{}
\author{Naoto Onda}{OMRON SINIC X Corporation, Tokyo, Japan}{naoto.onda@sinicx.com}{}{}

\authorrunning{S.~Sonoda et al.} 

\Copyright{Sho Sonoda, Kazumi Kasaura, Yuma Mizuno, Kei Tsukamoto, and Naoto Onda} 



\relatedversiondetails[]{Preprint Version}{https://arxiv.org/abs/2503.19605} 

\supplementdetails[linktext={GitHub repository}, subcategory={Source Code}]{Software}{https://github.com/auto-res/lean-rademacher/tree/f34ab4f0a9029682f1af3179a3fe9b5e114e511f}

\funding{This work was supported by 
JSPS KAKENHI 24K21316,
and JST Moonshot R\&D Program JPMJMS2236, PRESTO JPMJPR2125, BOOST JPMJBY24E2, and CREST JPMJCR25I5}

\acknowledgements{We thank the reviewers for useful feedback on our manuscript.}

\nolinenumbers

\EventEditors{Ekaterina Komendantskaya and Tobias Nipkow}
\EventNoEds{2}
\EventLongTitle{17th International Conference on Interactive Theorem Proving (ITP 2026)}
\EventShortTitle{ITP 2026}
\EventAcronym{ITP}
\EventYear{2026}
\EventDate{July 26--29, 2026}
\EventLocation{Lisbon, Portugal}
\EventLogo{}
\SeriesVolume{382}
\ArticleNo{8}

\begin{document}

\maketitle

\begin{abstract}
Understanding and certifying the generalization performance of machine learning algorithms---i.e. obtaining theoretical estimates of the test error from the training error---is a central theme of statistical learning theory.
Among the many complexity measures used to derive such guarantees, \emph{Rademacher complexity} yields sharp, data-dependent bounds that apply well beyond classical VC-dimension theory.
In this study, we formalize the generalization error bound by \emph{Rademacher complexity} in Lean~4, building on measure-theoretic probability theory available in the Mathlib library.
Our development provides a mechanically-checked pipeline from the definitions of empirical and expected Rademacher complexity, through a formal symmetrization argument and a bounded-differences analysis, to high-probability uniform deviation bounds via a formally proved McDiarmid inequality.
A key technical contribution is a reusable mechanism for lifting results from \emph{countable} hypothesis classes (where measurability of suprema is straightforward in Mathlib) to \emph{separable} topological index sets via a reduction to a countable dense subset.
As worked applications of the abstract theorem, we mechanize standard empirical Rademacher bounds for linear predictors under $\ell_2$ and $\ell_1$ regularizations, and we also formalize a Dudley-type entropy integral bound based on covering numbers and a chaining construction.
\end{abstract}

\section{Introduction}
Generalization analysis is the task of estimating, from finite data, how accurately a learned model will perform on \emph{unseen} test data.
Formally, let $\calX$ be a data domain and let $\mu$ be an unknown data-generating distribution on $\calX$.
A learning algorithm observes an i.i.d.\ sample $S=(x_1,\dots,x_n)$ drawn from $\mu$ and selects a hypothesis $h$ from some hypothesis class $\mathcal{H}$ (or, more specifically, a function from a parameterized family).
If $\ell(h,x)$ denotes a loss value on data $x$, the \emph{population risk} (test error) and the \emph{empirical risk} (training error) are defined as
\[
L(h) := \mathbb{E}_{x\sim\mu}[\ell(h,x)],
\qquad
\widehat L_S(h) := \frac1n\sum_{k=1}^n \ell(h,x_k).
\]
Here a ``data point'' may already contain both an input and a label.
For example, in supervised regression one may take $\calX=\mathcal{U}\times\mathbb{R}$, write a sample point as $x=(u,y)$, and let a predictor $h:\mathcal{U}\to\mathbb{R}$ have squared-error loss
\[
\ell(h,(u,y)) := (h(u)-y)^2.
\]
Then
\[
L(h)=\mathbb{E}_{(u,y)\sim\mu}\big[(h(u)-y)^2\big],
\qquad
\widehat L_S(h)=\frac1n\sum_{k=1}^n (h(u_k)-y_k)^2
\]
for a sample $S=((u_1,y_1),\ldots,(u_n,y_n))$.
A typical goal in statistical learning theory is to bound the \emph{generalization gap} $|\widehat L_S(h)-L(h)|$, uniformly over all hypotheses in the class (or at least over the hypotheses produced by a learning algorithm), with high probability over the sample draw.

In this study, we formalize the generalization error bound by \emph{Rademacher complexity} \cite{Bartlett2002} in Lean~4 \cite{deMouraUllrich2021lean4} based on probability theory formalized in Lean~4's mathematical library, Mathlib \cite{Mathlib}.
We refer to Mohri et al.\ (2018) \cite{Mohri2018} as a standard textbook on statistical machine learning theory, and Wainwright (2019) \cite{Wainwright2019} for more mathematical details on the Rademacher complexity and concentration inequalities used in generalization error analysis.
Concretely, rather than fixing a particular learning algorithm, we formalize a uniform law of large numbers (ULLN) style deviation functional, which we call the \emph{uniform deviation}: for a family of measurable real-valued functions $F=\{f_i:\calX\to\mathbb{R}\}_{i\in\iota}$ we consider
\[
\mathrm{UD}_n(F;S)
:=\sup_{i\in\iota}\left|\frac1n\sum_{k=1}^n f_i(x_k) - \mathbb{E}_{x\sim\mu}[f_i(x)]\right|.
\]
This quantity subsumes classical generalization-gap bounds by taking $f_i$ to be loss functions $\ell(h_i,\cdot)$, and it also aligns well with the measure-theoretic infrastructure in Mathlib.
In the Lean code, the corresponding objects are defined as \lstinline|uniformDeviation| and bounded using \lstinline|empiricalRademacherComplexity| and \lstinline|rademacherComplexity| (Section~4 of the main body).

\paragraph*{Why Rademacher complexity?}
Classical sample-complexity theory for the \emph{PAC learning} framework \cite{Valiant1984} often proceeds via the \emph{Vapnik--Chervonenkis (VC) dimension} \cite{Vapnik1971}.
While powerful, VC-dimension-based bounds are typically \emph{data-independent} and tailored to worst-case uniform guarantees for $0$--$1$ classification.
In modern learning settings, however, one often seeks bounds that (i) apply to real-valued losses and function classes beyond binary classifiers, and (ii) adapt to the \emph{observed sample} through a data-dependent complexity.
Rademacher complexity provides precisely such a tool: it measures the ability of a function class to correlate with random signs, and yields sharp uniform deviation bounds for a wide variety of learning problems, including analyses that underpin \emph{kernel methods} and \emph{deep learning} \cite{Mohri2018,Telgarsky2021note,Bach2024book}.
This flexibility motivates a foundational formalization effort: once the abstract Rademacher pipeline is mechanized, many concrete generalization analyses can be obtained as \emph{specializations} by instantiating the function class and bounding its complexity.

\paragraph*{A textbook-level bound and its formal counterpart.}
At a high level, the classical Rademacher approach yields the following pattern.
Assume a uniform bound $|f_i(x)|\le b$.
Then one first proves an \emph{expected} bound
\[
\mathbb{E}[\mathrm{UD}_n(F;S)] \le 2\,\mathfrak{R}_n(F),
\]
where $\mathfrak{R}_n(F)$ is the (distribution-dependent) Rademacher complexity of $F$.
Then one upgrades this to a \emph{high-probability} statement using a concentration inequality (typically McDiarmid's inequality), obtaining for all $\delta\in(0,1)$
\[
\mathbb{P}\left(\mathrm{UD}_n(F;S) \le 2\,\mathfrak{R}_n(F) + b\sqrt{\frac{2\log(1/\delta)}{n}}\right)\ge 1-\delta.
\]
Our Lean development follows and formalizes this standard ``symmetrization + bounded differences'' route, and exposes the corresponding countable-index and separable-index tail theorems in \lstinline|FoML/Main.lean|.

\paragraph*{Why Lean and Mathlib?}
A recurring theme in modern learning theory is that generalization bounds are most naturally expressed and proved in a measure-theoretic language: product measures for i.i.d.\ sampling, measurability of suprema, integrability of envelopes, and concentration inequalities stated for measurable functions on probability spaces.
Over the past several years, measure theory and probability theory in Lean's ecosystem have developed rapidly, and Mathlib has become a particularly convenient target for such work due to its extensive analysis libraries and the large community of mathematically oriented contributors.
These factors make Lean~4 + Mathlib a natural setting for formalizing generalization analysis at a level close to contemporary textbooks (as opposed to highly simplified, finite or discrete proxies).

\paragraph*{Novelty and contributions.}
Prior work has formalized generalization bounds, but only in relatively simplified settings.
For example, Bagnall and Stewart (2019) \cite{Bagnall2019gerrorCoq} formalize a generalization bound for neural networks, but restrict parameters to a finite set, which is insufficient for formalizing today’s diverse generalization analyses.
Our contribution is, to our knowledge, the \emph{first} systematic formalization of generalization analysis via Rademacher complexity.
The classical inequalities we use are mostly known on paper; the new mathematical clarification produced by the formalization is the precise package of sufficient conditions needed to make uncountable suprema measurable in the Rademacher pipeline.
In particular, the Lean proof isolates a reusable countable-dense reduction for separable index spaces, together with continuity-under-integration hypotheses such as first countability and uniform domination, conditions that are routinely implicit in empirical-process arguments but are not presented as a single textbook lemma.
More specifically, the repository provides:
\begin{itemize}
\item a reusable Lean formalization of empirical and expected Rademacher complexity and the associated uniform deviation functional;
\item a mechanically checked symmetrization argument connecting uniform deviations to Rademacher averages;
\item a formally verified bounded-differences (Lipschitz) analysis of the uniform deviation functional and a full McDiarmid inequality proof, yielding high-probability generalization bounds;
\item a principled extension from countable index sets to separable topological parameter spaces via a countable dense reduction, resolving measurability issues of uncountable suprema;
\item reusable formal infrastructure for maximal inequalities, Massart's lemma, PMF-based Rademacher variables, empirical pseudo-metrics, covering numbers, covering chains, and Dudley's entropy integral;
\item worked applications: formal generalization-relevant Rademacher bounds for $\ell_2$- and $\ell_1$-regularized linear predictors, and a formal Dudley entropy integral bound via covering numbers and chaining.
\end{itemize}

\paragraph*{Paper organization.}
Section~2 summarizes the repository and its measure-theoretic scope.
Sections~3--4 define the core objects and present the main Rademacher-based expectation and tail bounds.
Section~5 develops the $\ell_2$ and $\ell_1$ linear predictor examples.
Section~6 presents the Dudley entropy integral formalization.
Section~7 discusses key design choices and engineering challenges that arise when translating textbook arguments into Mathlib's formal framework, and Section~8 situates the contribution in the broader literature on formalized learning theory and concentration inequalities.
 \section{Repository Overview and Scope}

The repository’s core contribution is a mechanically-checked pipeline from (i) definitions of empirical and expected Rademacher complexity, through (ii) a formal symmetrization argument, to (iii) high-probability uniform deviation (ULLN-style) bounds via a formally proved McDiarmid inequality, with an additional reduction step that extends the results from countable hypothesis classes to separable topological index sets. It also includes worked examples showing how $\ell_2$- and $\ell_1$-type norm constraints yield standard $O(1/\sqrt{n})$ and $O(\sqrt{\log (d)/n})$ empirical Rademacher-complexity bounds for linear prediction, and a formal Dudley entropy-integral upper bound for empirical Rademacher complexity (via covering numbers and a chaining construction).

From a theorem proving perspective, the notable aspects are: the explicit handling of measurability/integrability hypotheses (often forcing \lstinline|Countable| index sets, or separability+continuity to reduce to a countable dense subclass), the use of product measures $\mu^n$ rather than i.i.d.\ sequences as primitive, and the systematic bridging between informal learning-theory arguments and Mathlib's measure-theoretic infrastructure. Conceptually, the main theorems align with classical results in Rademacher complexity and generalization theory as in Bartlett--Mendelson (risk bounds via Rademacher averages), McDiarmid's bounded-differences method, and Dudley's metric-entropy/chaining bounds. The formalization contribution is to make the side conditions, especially measurability of suprema and integrability under product measures, explicit enough to be reused by later learning-theory developments.

\paragraph*{Scope and assumptions.}

Our Lean development is structured around a general measure-theoretic formulation of
empirical processes:
a probability space $(\Omega,m\Omega,\mu)$, 
a data space $\mathcal{X}$ (sometimes \lstinline|Z| in code), 
and a random variable $X:\Omega\to\mathcal{X}$ (encoding the data distribution).
Fix an index type $\iota$ (parameterizing hypotheses), 
and a family of real-valued functions
\[
\calF \;=\; \{f_i : \calX\to\RR \}_{i\in\iota}.
\]
In the code, such a class is represented by a single curried term \lstinline{f : ι → 甲 → ℝ}. 

Samples of size $n$ are modeled as functions $\omega:\mathrm{Fin}\ n\to\Omega$, pushed forward through $X$ to obtain $S:=X\circ \omega:\mathrm{Fin}\ n\to\mathcal{X}$. Expectations are integrals in Mathlib notation, e.g.\ \lstinline|μ[g]| for $\int g\,d\mu$.

We assume the reader is comfortable with basic probability/analysis (product measures, integrals, concentration inequalities, norms) but not necessarily with ML-specific vocabulary (hypothesis classes, generalization error, empirical risk). We will use natural logarithms $\log$ (Lean’s \lstinline|Real.log|) and the convention that $n\in\mathbb{N}$, $n>0$ when dividing by $n$ or using $\sqrt{n}$. We interpret \lstinline|toReal| of an \lstinline|ENNReal| probability as the ordinary probability in $[0,1]$ (a standard Mathlib bridge).

\section{Core Definitions and Notational Conventions}
The central definitions in the file
\lstinline|FoML/Defs.lean| include the empirical and (expected) Rademacher
complexities:

\paragraph*{Rademacher signs as a finite type.}
The repository models a sign vector as a function \lstinline|Fin n → {-1,1}| in $\mathbb{Z}$, with \lstinline|Fintype| structure giving a finite uniform averaging over all $2^n$ sign assignments.

\begin{lstlisting}
-- FoML/Defs.lean
def Signs (n : ℕ) : Type := Fin n → ({-1, 1} : Finset ℤ)
\end{lstlisting}

Mathematically, one can view \lstinline|Signs n| as $\{\pm 1\}^n$ equipped with the uniform distribution; Lean often uses explicit finite sums multiplied by $(\#\mathrm{Signs}\ n)^{-1}$ instead of an expectation over i.i.d.\ $\sigma_k$.

\paragraph*{Empirical Rademacher complexity.}
For a sample $S:\mathrm{Fin}\ n\to\mathcal{X}$ and a class $f:\iota\to\mathcal{X}\to\mathbb{R}$, the code defines:
\[
\emprad_n(f;S)
:= \frac{1}{|\{\pm1\}^n|}\sum_{\sigma\in\{\pm1\}^n}
\sup_{i\in\iota}\left|\frac{1}{n}\sum_{k=1}^n \sigma_k\, f_i(S_k)\right|.
\]
This matches the common ML definition up to the choice of explicit averaging versus probabilistic expectation.

\begin{lstlisting}
-- FoML/Defs.lean
def empiricalRademacherComplexity (n : ℕ) (f : ι → 甲 → ℝ) (S : Fin n → 甲) : ℝ :=
  (Fintype.card (Signs n) : ℝ)⁻¹ *
    ∑ σ : Signs n, ⨆ i, |(n : ℝ)⁻¹ * ∑ k : Fin n, (σ k : ℝ) * f i (S k)|
\end{lstlisting}

\paragraph*{Expected Rademacher complexity.}
For a random variable $X:\Omega\to\mathcal{X}$ and probability measure $\mu$ on $\Omega$, the development forms the product measure $\mu^n$ on $\Omega^n$ and defines:
\[
\rad_n(f;\mu,X):=\mathbb{E}_{\omega\sim \mu^n}\big[\emprad_n(f;X\circ \omega)\big].
\]

\begin{lstlisting}
-- FoML/Defs.lean
def rademacherComplexity (n : ℕ) (f : ι → 甲 → ℝ) (μ : Measure Ω) (X : Ω → 甲) : ℝ :=
    μⁿ[fun ω : Fin n → Ω ↦ empiricalRademacherComplexity n f (X ∘ ω)]
\end{lstlisting}

\paragraph*{Uniform deviation.}
The central “generalization” quantity is the supremum of the absolute difference between the empirical mean and the population mean:
\[
\mathrm{UD}_n(f;\mu,X;S)
:=\sup_{i\in\iota}\left|\frac1n\sum_{k=1}^n f_i(S_k)\;-\;\mathbb{E}_\mu[f_i(X)]\right|.
\]

\begin{lstlisting}
-- FoML/Defs.lean
def uniformDeviation (n : ℕ) (f : ι → 甲 → ℝ) (μ : Measure Ω) (X : Ω → 甲) (S : Fin n → 甲) : ℝ :=
    ⨆ i, |(n : ℝ)⁻¹ * ∑ k : Fin n, f i (S k) - μ[fun ω' ↦ f i (X ω')]|
\end{lstlisting}

\paragraph*{One-sided Rademacher complexity.}
The repo also defines \lstinline|empiricalRademacherComplexity_without_abs|, a one-sided empirical Rademacher complexity obtained by dropping the absolute value inside the supremum.
This variant appears because Massart- and Dudley-style chaining arguments naturally bound a one-sided supremum of signed sums.
The absolute-value version remains the main quantity for uniform deviations; the development keeps both definitions explicit and proves comparison lemmas so that the exact assumptions needed to move between one-sided and absolute-value statements are visible.

\section{Generalization via Rademacher Complexity and Concentration}

\paragraph*{Symmetrization as a first-class formal object.}
The file \lstinline|FoML/Symmetrization.lean| develops an integral identity that replaces a supremum over differences of two independent samples by an average over sign-flipped sums. This is the backbone of standard Rademacher complexity arguments, and it is exposed as a theorem used downstream:
\lstinline|abs_symmetrization_equation|. The formal version is phrased over the product measure on $(\Omega\times\Omega)^n$ and uses finitary sign averaging.

\paragraph*{Expected supremum bound.}
In \lstinline|FoML/Rademacher.lean|, the key theorem \lstinline|expectation_le_rademacher| states (informally):
\[
\mathbb{E}_{\omega\sim\mu^n}\left[\sup_{i\in\iota}\left|\sum_{k=1}^n f_i(X(\omega_k)) - n\,\mathbb{E}_\mu[f_i(X)]\right|\right]
\;\le\;2n\cdot \rad_n(f;\mu,X),
\]
under \lstinline|Nonempty ι|, \lstinline|Countable ι|, measurability hypotheses, and a uniform bound $|f_i(z)|\le b$. The proof explicitly composes: (i) rewriting the mean as a coordinatewise mean on the product space, (ii) applying the symmetrization identity, and (iii) splitting the supremum via triangle inequality (a step corresponding to the classical “ghost sample + symmetrization $\Rightarrow 2\kern1pt\rad$” argument). For context, this is the standard route by which Rademacher complexity controls expected uniform deviations and risk bounds.

\begin{lstlisting}
-- FoML/Rademacher.lean (excerpt)
theorem expectation_le_rademacher {X : Ω → Z} [Nonempty ι] [Countable ι]
    (hf : ∀ i, Measurable (f i ∘ X))
    {b : ℝ} (hb : b ≥ 0) (hf' : ∀ (i : ι) (z : Z), |f i z| ≤ b) :
    μⁿ[fun ω : Fin n → Ω ↦ ⨆ i, |∑ k : Fin n, f i (X (ω k)) - n • μ[fun ω' ↦ f i (X ω')]|]
    ≤ (2 * n) • rademacherComplexity n f μ X := by ...
\end{lstlisting}

\paragraph*{Normalized form.}
The file \lstinline|FoML/Main.lean| packages a normalized statement for \lstinline|uniformDeviation|, dividing the previous bound by $n$:
\[
\mathbb{E}_{\omega\sim\mu^n}\big[\mathrm{UD}_n(f;\mu,X;X\circ\omega)\big]\;\le\;2\kern1pt\rad_n(f;\mu,X).
\]
This is the canonical “expected generalization gap bound” in Rademacher complexity theory.

\begin{lstlisting}
-- FoML/Main.lean (excerpt)
theorem uniform_deviation_expectation_le_two_smul_rademacher_complexity
    [Nonempty ι] [Countable ι] [IsProbabilityMeasure μ]
    (hn : 0 < n) (X : Ω → 甲)
    (hf : ∀ i, Measurable (f i ∘ X))
    {b : ℝ} (hb : 0 ≤ b) (hf' : ∀ i x, |f i x| ≤ b) :
    μⁿ[fun ω : Fin n → Ω ↦ uniformDeviation n f μ X (X ∘ ω)]
    ≤ 2 • rademacherComplexity n f μ X := by ...
\end{lstlisting}

\paragraph*{Bounded differences for samples.}
To obtain high-probability bounds, the repo proves that changing one sample point changes \lstinline|uniformDeviation| by at most $2b/n$ (the standard bounded differences constant) when all functions are bounded by $b$:
\[
\big|\mathrm{UD}_n(S)-\mathrm{UD}_n(S^{(i\leftarrow x')})\big|\le \frac{2b}{n}.
\]

\begin{lstlisting}
-- FoML/BoundedDifference.lean (excerpt)
theorem uniformDeviation_bounded_difference
    (hn : 0 < n) (X : Ω → 甲) (hf : ∀ i, Measurable (f i ∘ X))
    {b : ℝ} (hb : 0 ≤ b) (hf': ∀ i z, |f i z| ≤ b)
    (i : Fin n) (S : Fin n → 甲) (x' : 甲) :
    |uniformDeviation n f μ X S - uniformDeviation n f μ X (Function.update S i x')|
    ≤ (n : ℝ)⁻¹ * 2 * b := by ...
\end{lstlisting}

\paragraph*{McDiarmid tail bound and the countable theorem.}
Using a formal McDiarmid inequality (file \lstinline|FoML/McDiarmid.lean|), \lstinline|FoML/Main.lean| proves a tail bound around the mean and then combines it with the expectation bound to obtain the standard Rademacher-complexity tail estimate:
\[
\mathbb{P}\!\left(
\mathrm{UD}_n \;\ge\; 2\kern1pt\rad_n + \varepsilon
\right)
\;\le\;\exp\!\left(-\frac{\varepsilon^2\,n}{2b^2}\right),
\]
where $|f_i(x)|\le b$ uniformly. This is precisely the usual “symmetrization + McDiarmid” pathway found in standard textbooks and original papers on learning bounds.

\begin{lstlisting}
-- FoML/Main.lean (excerpt)
theorem uniform_deviation_tail_bound_countable_of_pos
    [MeasurableSpace 甲] [Nonempty 甲] [Nonempty ι] [Countable ι] [IsProbabilityMeasure μ]
    (f : ι → 甲 → ℝ) (hf : ∀ i, Measurable (f i))
    (X : Ω → 甲) (hX : Measurable X)
    {b : ℝ} (hb : 0 < b) (hf' : ∀ i x, |f i x| ≤ b)
    {ε : ℝ} (hε : 0 ≤ ε) :
    (μⁿ (fun ω ↦ 2 • rademacherComplexity n f μ X + ε ≤ uniformDeviation n f μ X (X ∘ ω))).toReal
    ≤ (- ε ^ 2 * n / (2 * b ^ 2)).exp := by ...   
\end{lstlisting}

\paragraph*{Extending from countable to separable index sets.}

A recurring obstacle in mechanizing learning theory is measurability of suprema: \texttt{mathlib} provides \lstinline{Measurable.iSup} for countable index types, but for uncountable parameter spaces measurability is not automatic. 
The development addresses this by separating \emph{proof obligations} from \emph{modeling convenience}. Theorem \lstinline{main_countable} is proved for \lstinline{[Countable ι]}; then \lstinline{main_separable} lifts the result to the common case where $\iota$ is a separable topological space and the map $i\mapsto f_i(x)$ is continuous for each $x$.

The lifting works by reducing $\sup_{i\in\iota}$ to a supremum over a countable dense sequence. Formally, \lstinline{FoML/SeparableSpaceSup.lean} proves a lemma (for real-valued continuous functions)
\[
\sup_{x\in X} g(x) \;=\; \sup_{n\in\NN} g(\mathrm{denseSeq}(n)),
\]
stated as \lstinline{separableSpaceSup_eq_real}. This lemma is then used to show definitional equivalence of the learning-theoretic quantities when replacing $f:\iota\to(\calX\to\RR)$ by its restriction along \lstinline{denseSeq ι : Nat → ι}. In particular, \lstinline{FoML/Main.lean} proves equalities
\[
\emprad_n(\calF\;x)
=
\emprad_n(\calF\circ \mathrm{denseSeq}\;x),
\qquad
\rad_n(\calF)
=\rad_n(\calF\circ \mathrm{denseSeq}),
\qquad
\Delta_n
=\Delta_n^{\mathrm{denseSeq}},
\]
These are packaged as equality lemmas for empirical complexity, expected complexity, and uniform deviation.
The last equality is technically the most subtle: it requires continuity of the integral term $i\mapsto\int f_i(X(\omega)),d\mu(\omega)$, which is obtained via a dominated-continuity lemma under the uniform bound $|f_i(x)|\le b$.

The final theorem \lstinline{main_separable} applies \lstinline{main_countable} to the countable family \lstinline{f ∘ denseSeq ι} and rewrites the result back using the aforementioned equalities. Mathematically, it yields the same tail bound as \lstinline{main_countable} but under topological assumptions (\lstinline{[SeparableSpace ι]} plus continuity) rather than the purely set-theoretic \lstinline{[Countable ι]} assumption.

\begin{lstlisting}
-- FoML/Main.lean (excerpt)
theorem uniform_deviation_tail_bound_separable_of_pos
    [MeasurableSpace 甲] [Nonempty 甲] [Nonempty ι]
    [TopologicalSpace ι] [SeparableSpace ι] [FirstCountableTopology ι]
    [IsProbabilityMeasure μ]
    (f : ι → 甲 → ℝ) (hf : ∀ i, Measurable (f i))
    (X : Ω → 甲) (hX : Measurable X)
    {b : ℝ} (hb : 0 < b) (hf' : ∀ i x, |f i x| ≤ b)
    (hf'' : ∀ x : 甲, Continuous fun i ↦ f i x)
    {ε : ℝ} (hε : 0 ≤ ε) :
    (μⁿ (fun ω ↦ 2 • rademacherComplexity n f μ X + ε ≤ uniformDeviation n f μ X (X ∘ ω))).toReal
    ≤ (- ε ^ 2 * n / (2 * b ^ 2)).exp := by ...
\end{lstlisting}

\section{Linear Predictors with \texorpdfstring{$\ell_2$ and $\ell_1$}{ell2 and ell1} Regularization}
To demonstrate the power of the abstract infrastructure, we specialize the
Rademacher complexity to linear hypothesis classes with norm constraints.
These are the simplest continuous model classes used in learning theory: a parameter vector $w$ defines a prediction rule $x\mapsto\langle w,x\rangle$, and different norm constraints on $w$ capture different geometric assumptions on the model class.

\paragraph*{\texorpdfstring{$\ell_2$}{ell2}-bounded predictors.}
Let $d\in\mathbb{N}$ and consider the Euclidean space $\mathbb{R}^d$ (modeled as \lstinline|EuclideanSpace ℝ (Fin d)|). For weights $w$ in an $\ell_2$ ball of radius $W$ and inputs $x$ in an $\ell_2$ ball of radius $X$, the class is
\[
f_w(x)=\langle w,x\rangle.
\]
The repo formalizes the standard empirical bound:
\[
\emprad_n(\{f_w:\|w\|_2\le W\}; S)\;\le\;\frac{XW}{\sqrt{n}}.
\]
The proof is recognizably the classical argument: rewrite the supremum via Cauchy--Schwarz, then compute the second moment of the Rademacher-weighted sum using sign orthogonality (all cross terms vanish). The orthogonality lemma is proved by an explicit sign-flip involution (\lstinline|rademacher_flip|) in \lstinline|FoML/RademacherVariableProperty.lean|.

\begin{lstlisting}
-- FoML/Main.lean (excerpt)
theorem linear_predictor_l2_bound
    [Nonempty ι] (d : ℕ) (W X : ℝ) (hx : 0 ≤ X) (hw : 0 ≤ W)
    (Y' : Fin n → Metric.closedBall (0 : EuclideanSpace ℝ (Fin d)) X)
    (w' : ι → Metric.closedBall (0 : EuclideanSpace ℝ (Fin d)) W) :
    empiricalRademacherComplexity
      n (fun (i : ι) a ↦ ⟪((Subtype.val ∘ w') i), a⟫) (Subtype.val ∘ Y')
    ≤ X * W / √(n : ℝ) := by ...
\end{lstlisting}

\paragraph*{\texorpdfstring{$\ell_1$}{ell1}-bounded predictors.}
For $\ell_1$-bounded weights $\|w\|_1\le W$ and $\ell_\infty$-bounded inputs $\|x\|_\infty\le X_\infty$, standard learning theory gives
\[
\emprad_n(\{x\mapsto \langle w,x\rangle:\|w\|_1\le W\}; S)
\;\lesssim\;\frac{X_\infty W}{\sqrt{n}}\sqrt{2\log(2d)}.
\]
The repo proves essentially this statement (with explicit constants), using a Massart-type finite-class lemma for a $2d$-sized auxiliary class corresponding to signed coordinate projections; and a duality inequality reflecting $\sup_{\|w\|_1\le W}\langle w,z\rangle = W\|z\|_\infty$ (implemented with careful absolute values and finiteness). Classical references typically attribute the finite-class step to Massart and standard concentration methods.

\begin{lstlisting}
-- FoML/Main.lean (excerpt)
theorem linear_predictor_l1_bound
    [Nonempty ι] (d : ℕ) (Xinf W : ℝ)
    (hX : 0 ≤ Xinf) (hW : 0 ≤ W) (d_pos : 0 < d) (n_pos : 0 < n)
    (Y' : Fin n → LinftyBall (d := d) Xinf)
    (w' : ι → L1Ball (d := d) W) :
    empiricalRademacherComplexity n
      (fun i a => (∑ j : Fin d, (w' i).1 j * a j))
      (Subtype.val ∘ Y')
    ≤ (Xinf * W / Real.sqrt (n : ℝ)) * Real.sqrt (2 * Real.log (2 * d)) := by ...
\end{lstlisting}

\section{Dudley Entropy Integral Bound}
Beyond linear predictors, we formalized the Dudley entropy integral bound, which
relates the Rademacher complexity of an arbitrary function class to its covering
numbers under the empirical $L^2$ metric.
Here ``entropy'' means the logarithm of a covering number: it measures how many functions are needed to approximate the whole class at scale $\varepsilon$ on the observed sample.
This required a substantial additional infrastructure layer, because Dudley's theorem is naturally expressed in terms of metric entropy with respect to a sample-dependent pseudometric.

\paragraph*{Empirical pseudo-metric and covering numbers.}
For a fixed sample $S:\mathrm{Fin}\ n\to\mathcal{X}$ and functions $f,g:\mathcal{X}\to\mathbb{R}$, the repo defines an empirical $\ell_2$-type seminorm
\[
\|f-g\|_S := \sqrt{\frac1n\sum_{k=1}^n (f(S_k)-g(S_k))^2},
\]
implemented via \lstinline|empiricalNorm| and \lstinline|empiricalDist| in \lstinline|FoML/PseudoMetric.lean|. This induces a pseudo-metric structure on the set of functions restricted to the sample.

Covering numbers are then defined for totally bounded sets in a pseudo-metric space:
\[
N(\varepsilon) := \min\left\{|T| : T\subseteq A,\ A\subseteq \bigcup_{t\in T}B(t,\varepsilon)\right\},
\]
implemented as \lstinline|coveringNumber ha ε| in \lstinline|FoML/CoveringNumber.lean|.
The Lean term uses \lstinline|Nat.find|, Lean's operator for extracting the least natural number satisfying a proven existential statement; here the existential statement is the finite-cover lemma for totally bounded sets.

\paragraph*{The Dudley-type theorem as formalized.}
The central statement in \lstinline|FoML/Main.lean| is a wrapper around \lstinline|dudley_entropy_integral'| from \lstinline|FoML/DudleyEntropy.lean|. It bounds the \emph{without-absolute-value} empirical Rademacher complexity in terms of an entropy integral:
\[
\emprad^{(\mathrm{no\ abs})}_n(F;S)\;\le\;
4\varepsilon \;+\; \frac{12}{\sqrt{n}}
\int_{\varepsilon}^{c/2} \sqrt{\log N(u)}\,du,
\]
where $c$ is an upper bound on all empirical norms $\|F(f)\|_S$ and $\varepsilon<c/2$, and where $N(u)$ is the covering number of the class realized as a subset of \lstinline|EmpiricalFunctionSpace|.

\begin{lstlisting}
-- FoML/Main.lean (excerpt)
theorem dudley_entropy_integral_bound
  {甲 : Type v} {n : ℕ} {ι : Type u} [Nonempty ι] {F : ι → 甲 → ℝ} {S : Fin n → 甲} {c ε : ℝ}
  (ε_pos : 0 < ε) (h' : TotallyBounded (Set.univ : Set (EmpiricalFunctionSpace F S)))
  (m_pos : 0 < n) (cs : ∀ f : ι, empiricalNorm S (F f) ≤ c)
  (ε_le_c_div_2 : ε < c/2) :
    empiricalRademacherComplexity_without_abs n F S ≤
    (4 * ε + (12 / Real.sqrt n) *
    (∫ (x : ℝ) in ε..(c/2), √(Real.log (coveringNumber h' x)))) := by ...
\end{lstlisting}

\paragraph*{Relation to classical Dudley bounds.}
Dudley’s 1967 work \cite{Dudley1967} introduced entropy integrals to control sample regularity and suprema of Gaussian processes.
Modern empirical-process theory adapts similar ``chaining'' ideas to Rademacher/Gaussian complexities, yielding inequalities of the form
\[
\mathbb{E}_{\sigma}\left[\sup_{f\in\mathcal{F}} \sum_i \sigma_i f(x_i)\right]\lesssim \int \sqrt{\log N(\varepsilon)}\,d\varepsilon.
\]
The Lean development follows this blueprint explicitly: it constructs dyadic radii $\varepsilon_j=c/2^j$, chooses finite $\varepsilon_j$-covers (via \lstinline|coveringFinset|), defines a chaining approximation \lstinline|chainApprox|, proves a telescoping decomposition, bounds the “tail” by $\varepsilon$ (Part~A), bounds increments by a Massart finite-class lemma (Part~B), and converts the resulting finite sum into an interval integral using a monotonicity-based Riemann-sum comparison.

 \section{Design Choices}

We review the particularly challenging components of the development and the methodological choices that enabled progress.

\paragraph*{Independent variables.}

In the Rademacher-complexity argument, the training dataset is assumed to be i.i.d. This is essential, for example, in the symmetrization argument.
In a textbook-style formulation, i.i.d. variables $X_1,\ldots,X_n$ are formulated as a map  $\mathrm{Fin}\,n \to \Omega \to \calX$.
However, we noticed that it is more convenient to formalize them as compositions of a single variable $X:\Omega\to\calX$ with the coordinate projections $\Omega^n\to\Omega$, taking $\Omega^n$ to be the base probability space.
In this view, independence need not be explicitly assumed as it follows directly from the construction,
which simplifies descriptions of theorems.

On the other hand, in McDiarmid's inequality, the random variables $X_1,\ldots,X_n$ are assumed to be independent but need not be identically distributed.
So we formalize them as distinct functions $X:\mathrm{Fin}\,n\to\Omega\to\RR$ and explicitly assume independence as an explicit condition. Here the base probability space is $\Omega$.

The fact that the former construction $\Omega^n \to \Omega \to \calX$ satisfies the independence condition required in the latter is a theorem that should be proved. We could not find this result in Mathlib, so we supplied our own proof.
We also attempted another formalization that defines Rademacher complexity via probability mass functions, but deriving its properties along this path proved difficult, and we abandoned that approach.

\paragraph*{Topological details of the index set.}
The Rademacher complexity is an expectation of the \emph{supremum} over a hypothesis class $\calF$,
and in modern machine learning settings, $\calF$ is often \emph{uncountable}.
When $\calF$ is uncountable, however, measurability is not generally preserved under pointwise $\sup$.
We therefore first prove results in the \emph{countable} case and then extend to the \emph{separable} case.
Even when uncountable, a separable family allows the $\sup$ of continuous functions to be computed over a countable dense subset, reducing to the countable case.
The formalization clarified the exact hypotheses needed for this reduction in the learning-theoretic setting: continuity must also be preserved after taking the population integral, which we prove using uniform domination and a first-countability assumption.
This package of sufficient conditions is routine in applications, but we were unable to locate it as a ready-made statement in standard learning-theory or empirical-process texts.

\paragraph*{Conditional expectation.}
Textbook proofs of McDiarmid's inequality typically proceed via conditional expectations.
However, because conditional expectation is defined abstractly, it is difficult to carry out concrete computations directly from that definition. 
(At the time we were developing the formalization, Mathlib had not yet included the Doob--Dynkin lemma; now there is \lstinline{Mathlib.MeasureTheory.Function.FactorsThrough}.)
We therefore avoided conditional expectations and instead defined the relevant quantities directly by integration, using independence. This changes the proof order: in the original argument the constructed sequence $Y$ is a martingale by construction, whereas in our approach (without conditional expectations) we establish the martingale property from independence.

\paragraph*{Integral and supremum.} A large fraction of the lines in many lemmas is devoted to handling integrability conditions (and the treatment of $\sup$), which are often considered routine. Each time we performed an algebraic manipulation under the integral sign, we had to supply a fresh proof of integrability.

\paragraph*{Dudley entropy integral and \lstinline|without_abs|.}
The Dudley entropy integral bound is stated for \lstinline|empiricalRademacherComplexity_without_abs| because the chaining proof controls a one-sided supremum of signed increments.
In standard learning-theory presentations, this distinction is often hidden by working with centered or symmetric classes, or by applying the argument to both a class and its negation.
Our Lean development keeps the one-sided and absolute-value versions separate, and records comparison lemmas in \lstinline|FoML/RademacherVariableProperty.lean| so that later users can see exactly which symmetry or boundedness assumptions are being used.

\paragraph*{Generative AI assistance.}
Our project spans more than two years of manual formalization.  During the early
stages, available large language models tended to produce Lean~3 code and were
not practically useful.  Since mid-2025 the models improved markedly and began
producing syntactically correct Lean~4 code; we used generative AI to assist
with routine yet tedious tasks such as rewriting expressions to fit Lean's type
system, managing casts between numeric types, and expanding algebraic identities.
For example, writing the empirical Rademacher complexity as a measure-theoretic
expectation required meticulous adjustments of types.  While generative AI
streamlined these technical steps, it could not reliably construct full proofs;
we therefore designed the high-level strategy and wrote most proofs manually.
 \section{Literature Overview}

\cref{tab:ml,tab:prob} summarize close literature and related formalizations used as context for this study.

\subsection{PAC Learning and VC Dimension (Lean, Rocq)}
\begin{table*}[t]
\centering
\caption{Formalization of machine-learning theory and related probabilistic analyses}
\label{tab:ml}
\begin{tabular}{@{}>{\raggedright\arraybackslash}p{.34\textwidth}>{\raggedright\arraybackslash}p{.14\textwidth}>{\raggedright\arraybackslash}p{.45\textwidth}@{}}
\toprule
Bentkamp et al. (2016,2019) \cite{Bentkamp2016deep-afp,Bentkamp2019expressive-dl} & Isabelle/HOL & Expressive power superiority of deep over shallow \\
Tassarotti and Harper (2018) \cite{Tassarotti2018itp} & Coq & Verified tail bounds for randomized programs via Karp-style recurrences \\
Bagnall and Stewart (2019) \cite{Bagnall2019gerrorCoq} & Coq & Generalization error bounds for finite hypothesis class\\
Tassarotti et al. (2021) \cite{Tassarotti2021pac} & Lean & PAC learnability of decision stumps\\
Vajjha et al. (2021) \cite{Vajjha2021certRL} & Coq & Convergence of reinforcement learning algorithms \\
Vajjha et al. (2022) \cite{Vajjha2022sgd} & Coq & Stochastic approximation theorem \\

Hirata (2025) \cite{hirata2025nofreelunch} & Isabelle/HOL & No free lunch theorem\\
(\emb{ours}) & Lean & Generalization error bound by Rademacher complexity \\ &&and Dudley's entropy integral\\
Zhang et al. (2026) \cite{zhang2026slt-lean4} & Lean & SLT with Gaussian Lipschitz concentration \\&&and Dudley’s entropy integral\\
\bottomrule
\end{tabular}
\end{table*}
 \begin{table*}[ht]
\centering
\caption{Formalization of Concentration Inequalities}
\label{tab:prob}
\begin{tabular}{@{}>{\raggedright\arraybackslash}p{.28\textwidth}>{\raggedright\arraybackslash}p{.65\textwidth}@{}}
\toprule
Markov/Chebyshev & Lean (Mathlib), Rocq (MathComp-Analysis\cite{MathComp-Analysis}, IBM/FormalML\cite{Vajjha2021certRL}),\\
& Isabelle/HOL (HOL-Probability)\\
Weak/strong laws of large numbers & Isabelle/HOL (AFP) \cite{Eberl2021lln-afp}\\
Azuma--Hoeffding & Lean (Mathlib) \\
McDiarmid & (\emb{ours}), Isabelle/HOL (AFP) \cite{Karayel2023concentration-afp}\\
Gaussian Lipschitz & Lean \cite{zhang2026slt-lean4}\\
\bottomrule
\end{tabular}
\end{table*}
 
A hypothesis class with finite VC dimension (or simply a finite class) admits Probably Approximately Correct (PAC) generalization
bounds. Early formalizations focused on specific cases as follows: 

\paragraph*{Finite Hypothesis Classes:} Bagnall \& Stewart (2019) \cite{Bagnall2019gerrorCoq} proved a general PAC
bound in \emph{Rocq} for any finite hypothesis class using Hoeffding's
inequality. Essentially, if \(\calF\) is finite, with probability
\(1-\delta\) the true error is within \(\eps\) of the training error
for
\(n \gtrsim \frac{1}{\eps^2}(\ln |\calF| + \ln\frac{1}{\delta})\).
Their Rocq development (part of the \emph{MLCERT} system) used a union
bound and Chernoff/Hoeffding bounds to link training and test
errors.
This formal result was applied to certify small neural network models' performance. However, it was limited to a finite set \(\calF\), excluding most modern model classes because $\calF$ with finite-dimensional parameters is an infinite set (so $|\calF| = \infty$).

\paragraph*{Decision Stump Class ($VC = 1$):} Tassarotti et al. (2021) \cite{Tassarotti2021pac} gave a full \emph{Lean 3} proof that the concept class of \emph{decision
stumps} (threshold classifiers in \(\RR\)) is
PAC-learnable. This is a classic textbook example with VC dimension 1. The formal proof uncovered subtle measure-theoretic issues that are glossed over in informal proofs. For instance, textbooks often assume measurability of argmax operations on sample data without proof---the formalization had to rigorously prove measurability and proper probability space definitions for the learning algorithm. The authors structured the proof to separate combinatorial reasoning about the algorithm's behavior from the analytic reasoning about probabilities. They employed the Giry monad (in Lean's category theory library) to handle distributions, and ultimately derived the standard PAC guarantee for decision stumps.

\paragraph*{Verified Tail Bounds for Randomized Programs:}
Tassarotti and Harper (2018) \cite{Tassarotti2018itp} mechanized Karp's theorem and extensions for randomized divide-and-conquer recurrences in Coq.
Their case studies include sequential and parallel QuickSort, random binary-search-tree height, and randomized leader election, with concrete numerical tail bounds.
Although this work is about randomized algorithms rather than statistical learning theory, it is related methodologically: it shows how nontrivial tail-bound arguments can be packaged into reusable theorem-proving infrastructure, albeit in a mainly discrete-probability setting rather than the measure-theoretic empirical-process setting used here.

\subsection{Formalizing Machine Learning Theory}

Statistical learning theory extends beyond classical generalization bounds. 

\paragraph*{Expressiveness of Deep Neural Networks:} 
Bentkamp et al. (2016,2019) \cite{Bentkamp2016deep-afp,Bentkamp2019expressive-dl}
formalized in \emph{Isabelle/HOL} a theorem that deep networks can represent certain functions exponentially more efficiently than shallow ones. Their formalization simplified and generalized the original proof within Isabelle's logic.
To support the proof, Bentkamp developed libraries for linear algebra
(matrix ranks), multivariate polynomials, and even Lebesgue measure
integration. The result is not about generalization error, but rather
about the capacity/representation power of deep vs. shallow
networks---nevertheless, it showcases the application of proof
assistants to core theoretical ML questions. It also enriched Isabelle's
libraries with notions like tensor products and rank, which are useful
in learning theory.

\paragraph*{Convergence of Reinforcement Learning:} Vajjha et
al. (2021) \cite{Vajjha2021certRL} verified the convergence of value iteration and policy iteration for Markov Decision Processes in \emph{Rocq}. In a follow-up, they formalized a stochastic approximation theorem \cite{Vajjha2022sgd} useful for
analyzing RL algorithms. These efforts, under the IBM FormalML
project, build a bridge between learning theory and formal verification
by proving probabilistic convergence properties in Rocq. They required
heavy use of Rocq's analysis libraries and bespoke techniques
(e.g.~coinduction for probabilistic processes).

\paragraph*{Statistical Learning Theory Based on Empirical Processes:} 

In early 2026, Zhang et al.~\cite{zhang2026slt-lean4} open-sourced an AI-assisted formalization of
statistical learning theory in Lean~4, focusing on empirical process theory.
They describe their generated development as being built from scratch.
At the level of initial seeds, a small amount of existing Lean code from developments such as \texttt{lean-rademacher}\footnote{\url{https://github.com/auto-res/lean-rademacher.git}} and \texttt{CLT}\footnote{\url{https://github.com/RemyDegenne/CLT.git}} was also present.
Given the scale of the generated library and its different goals, this is nevertheless best viewed as an independent, large-scale development.
Their system automatically generated a large number of definitions and proofs including Gaussian Lipschitz concentration and Dudley entropy integral for sub-Gaussian processes.

\subsection{Concentration Inequalities and Background Formalizations}

Many learning-theoretic proofs rely on \emph{concentration of measure}
results and related probabilistic inequalities. Over the 2010s--2020s,
these foundational results have been increasingly formalized, often as
prerequisites for the theorems above:

  \paragraph*{Isabelle/HOL:} Its probability theory library
  (\emph{HOL-Probability}) already included basic inequalities like
  \emph{Markov's inequality}, \emph{Chebyshev's inequality}, and
  exponential tail (Chernoff/Hoeffding) bounds by the late
  2010s.
  In 2023, Karayel and Tan \cite{Karayel2023concentration-afp} contributed an AFP entry
  \emph{``Concentration Inequalities''} which adds more advanced
  results. This includes \emph{Bennett's and Bernstein's inequalities}
  (for sub-exponential random variables), \emph{Efron--Stein's
  inequality} (variance bound via variance decomposition),
  \emph{McDiarmid's inequality} (bounded differences), and the
  \emph{Paley--Zygmund inequality}. Isabelle also has background limit
  theorems relevant to statistical reasoning: Eberl's AFP entry on the
  laws of large numbers \cite{Eberl2021lln-afp} proves the strong law
  for i.i.d.\ random variables and derives the weak law from it, using
  ergodic-theoretic infrastructure.
  Thus Isabelle/HOL provides a rich library of probability results.
  For the present Lean development, however, the goal was not merely to
  reproduce McDiarmid's inequality: the Rademacher-complexity pipeline
  also requires product-measure i.i.d.\ samples, measurability of uniform
  deviations, the countable-to-separable supremum reduction, and links to
  Mathlib's integration and topology APIs. The Isabelle results therefore
  served primarily as external sanity checks for the shape of statements,
  while the missing Lean-side infrastructure still had to be developed.

  \paragraph*{Lean:} Lean's Mathlib library \cite{Mathlib} gained basic measure theory around 2019,
  including Lebesgue integration and statistical independence. By mid-2020s, Mathlib
  had formal proofs of \emph{Markov and Chebyshev inequalities}. 
  This study also introduced
  \emph{Hoeffding's lemma} (a result that a bounded zero-mean variable is sub-Gaussian) and then proved \emph{Hoeffding's inequality} as a corollary of Chernoff bound techniques. It also implemented \emph{McDiarmid's inequality} from scratch in Lean 4.
  In 2025, Degenne implemented \emph{Azuma--Hoeffding} on Mathlib independently of our work.
  In 2026, Zhang et al.~\cite{zhang2026slt-lean4} implemented \emph{Dudley's entropy integral} (with other concentration inequalities such as \emph{Gaussian Lipschitz}) independently of our work.
  
  \paragraph*{Rocq:} Historically, Rocq's standard library did not include measure-theoretic concentration results.
  But developments like \emph{MathComp-Analysis} \cite{MathComp-Analysis} have added some
  pieces. 
Using these, the IBM FormalML team \cite{Vajjha2021certRL,Vajjha2022sgd} have formalized basic inequalities such as Markov and Chebyshev in Rocq.
  Hoeffding's and McDiarmid's inequalities were not present in Rocq as of early 2020s
  except in special-case proofs (e.g.~MLCert implicitly used a form of Hoeffding's bound for finite samples). 
  Tassarotti and Harper's Coq development \cite{Tassarotti2018itp} provides another related point in the design space: it verifies tail bounds for randomized programs using discrete probability and reusable recurrence theorems.
  In 2025, Affeldt et al.~\cite{affeldt2025itp-concentration-rocq} formalized concentration inequalities in the Rocq proof assistant, developing infrastructure for measure-theoretic reasoning and exponential bounds.  Their work includes Hoeffding's and Bernstein's inequalities. 

\section{Conclusion}
We presented a Lean~4 formalization of generalization analysis via Rademacher complexity, built on Mathlib's measure-theoretic probability infrastructure.
At the core is a mechanically checked pipeline that mirrors standard textbook proofs: (i) formal definitions of empirical and expected Rademacher complexity and of the uniform deviation functional, (ii) a symmetrization theorem that turns uniform deviations into sign-averaged suprema, and (iii) a bounded-differences argument combined with a fully formal McDiarmid inequality to obtain high-probability generalization bounds.
A major methodological contribution is the systematic treatment of measurability obstacles: results are first established for countable hypothesis classes (where measurability of suprema is readily available), and then lifted to separable topological parameter spaces through a countable dense reduction, together with continuity-under-integration lemmas.
This does not change the classical learning-theory bounds, but it clarifies a reusable set of sufficient conditions for the uncountable suprema that are routinely used in empirical-process arguments.

Beyond the abstract theorem, we demonstrated that this infrastructure supports nontrivial specializations.
We formalized classical empirical Rademacher bounds for linear predictors under $\ell_2$ and $\ell_1$ norm constraints, and we mechanized a Dudley entropy integral bound using covering numbers and a chaining construction.
These examples illustrate that once the general Rademacher pipeline is established in a proof assistant, many concrete bounds become ``executable corollaries'' obtained by instantiating the abstract objects and discharging model-specific analytic estimates.

Looking forward, we expect several extensions to be both practically useful and methodologically interesting: incorporating standard contraction inequalities for Lipschitz losses, connecting Rademacher complexity to Gaussian complexity and sub-Gaussian process tools, expanding libraries for covering-number estimates for common model classes, and improving automation for routine measurability and integrability obligations.
More broadly, we hope this development contributes to a foundation in which modern learning-theoretic proofs can be stated and verified at a faithful, measure-theoretic level, enabling trustworthy reuse in formal verification and certified ML pipelines.

\bibliography{main-itp2026}
\end{document}